\documentclass[letterpaper]{article} 
\usepackage{aaai24}  
\usepackage{times}  
\usepackage{helvet}  
\usepackage{courier}  
\usepackage[hyphens]{url}  
\usepackage{graphicx} 
\usepackage{natbib}  
\usepackage{caption} 
\usepackage{algorithm}
\usepackage{algorithmic}
\usepackage{booktabs}       
\usepackage{makecell}       
\usepackage{multirow}       
\usepackage{array}          
\usepackage{graphicx}       
\usepackage{caption}        
\usepackage{siunitx}        
\usepackage{amsfonts}
\usepackage{xcolor}
\usepackage[table]{xcolor}

\definecolor{lightblue}{RGB}{225, 240, 255} 
\definecolor{lightgray}{RGB}{235, 235, 235}     
\usepackage{newfloat}
\usepackage{listings}
\DeclareCaptionStyle{ruled}{labelfont=normalfont,labelsep=colon,strut=off} 
\floatstyle{ruled}
\newfloat{listing}{tb}{lst}{}
\floatname{listing}{Listing}
\usepackage{amsmath}

\title{
Knowledge-Guided Masked Autoencoder with Linear Spectral Mixing and Spectral-Angle–Aware Reconstruction}

\author{
    Abdul Matin\textsuperscript{\rm 1}, 
    Rupasree Dey\textsuperscript{\rm 1}, 
    Tanjim Bin Faruk\textsuperscript{\rm 1}, 
    Shrideep Pallickara\textsuperscript{\rm 1}, 
    Sangmi Lee Pallickara\textsuperscript{\rm 1}
}
\affiliations{

    \textsuperscript{\rm 1}Department of Computer Science, Colorado State University\\
    Fort Collins, CO, USA\\
    amatin@colostate.edu, rupasree.dey@colostate.edu, tanjim@colostate.edu, 
    shrideep@colostate.edu, sangmi@colostate.edu

}

\usepackage{bibentry}

\copyrighttext{This paper was accepted to the KGML Bridge at AAAI 2026 (non-archival).}

\begin{document}
\maketitle

\begin{abstract}
Integrating domain knowledge into deep learning has emerged as a promising direction for improving model interpretability, generalization, and data efficiency. In this work, we present a novel knowledge-guided ViT based Masked Autoencoder that embeds scientific domain knowledge within the self-supervised reconstruction process. Instead of relying solely on data-driven optimization, our proposed approach incorporates the Linear Spectral Mixing Model (LSMM) as a physical constraint and physically-based Spectral Angle Mapper (SAM) ensuring that learned representations adhere to known structural relationships between observed signals and their latent components. The framework jointly optimizes LSMM and SAM loss with a conventional Huber loss objective, promoting both numerical accuracy and geometric consistency in the feature space. This knowledge-guided design enhances reconstruction fidelity, stabilizes training under limited supervision, and yields interpretable latent representations grounded in physical principles. The experimental findings indicate that the proposed model substantially enhances reconstruction quality and improves downstream task performance, highlighting the promise of embedding physics-informed inductive biases within transformer-based self-supervised learning.
\end{abstract}

\section{Introduction}
\label{sec:intro}

Recent advances in self-supervised learning and large-scale transformer architectures have demonstrated impressive representation learning capabilities across vision and multimodal tasks. In particular, Vision Transformer-based Masked Autoencoders (ViT-MAEs) have shown that reconstructing masked input patches provides an effective pretext task for learning transferable features at scale \cite{he2022masked,dosovitskiy2021vit}. Such architectures have further evolved into large-scale \textit{foundation models} that can adapt across diverse visual and geospatial applications by leveraging vast unlabeled data \cite{bommasani2021foundation}. However, purely data-driven reconstruction objectives often overlook domain-specific priors and the physical mechanisms that govern data generation, limiting interpretability and generalization \cite{willard2020integrating,raissi2019pinn}.

In the context of Earth observation, hyperspectral satellite imagery offers a unique and information-rich modality. By capturing hundreds of contiguous spectral bands, these data provide a fine-grained description of surface materials, enabling precise detection of land-cover types, vegetation health, mineral composition, and soil properties \cite{zhang2023hyperspectral_review}. Despite their potential, hyperspectral datasets present distinct challenges for deep learning models: high spectral dimensionality, spatial redundancy, and the physical phenomenon of mixed pixels where each observation represents a linear or nonlinear combination of multiple materials. Consequently, the success of general-purpose foundation models cannot be directly transferred to hyperspectral data without incorporating the underlying physical knowledge that explains such mixtures.

In recent years, an emerging paradigm known as \emph{knowledge-guided machine learning} (KGML) has begun to blur the boundary between purely data-driven methods and science-driven modeling. Rather than treating machine learning as a black box, KGML frameworks strategically integrate domain knowledge (such as physical laws, mixing models, or causally grounded relationships) with representation learning in neural architectures \cite{karpatne2024kgmlsurvey}. For example, Liu et al. (2024) demonstrate how embedding process-based and remote-sensing knowledge into neural frameworks significantly improves soil-carbon estimation across agroecosystems, highlighting the potential of hybrid scientific–data approaches. Drawing on these insights, our work situates itself at the intersection of foundation-model design and domain-informed representation learning.

Motivated by these observations, we propose the \textbf{K}nowledge-\textbf{A}ugmented \textbf{R}econstruction with \textbf{M}asked \textbf{A}uto\textbf{e}ncoding (\textbf{\textsc{karma}}), a physics-informed ViT-MAE framework that fuses a physically interpretable model with transformer-based self-supervision. \textsc{karma} integrates the Linear Spectral Mixing Model (LSMM)~\cite{bioucas2012overview} as an inductive bias inside the decoder: the decoder predicts an abundance vector per patch, and the model reconstructs the observed signal as a linear combination of surface material signatures. This physics-guided structure ensures that learned latent representations remain consistent with real-world spectral composition laws, thereby promoting explainability and stability.

To further promote spectral fidelity and geometric alignment in the learned features, we jointly optimize a Spectral-Angle–aware loss along with Huber Loss and an explicit LSMM physics consistency loss. This hybrid objective encourages numerical accuracy, shape-preserving reconstructions, and explicit physical consistency which results in producing interpretable latent components that are useful for downstream tasks such as regression and classification. By embedding physical priors into a ViT-MAE backbone, \textsc{karma} bridges the gap between generic self-supervised learning and domain-grounded scientific understanding—offering a step toward foundation models that learn not only from data but also from the governing physical principles.

\noindent Our contributions include the following:
\begin{itemize}
  \item We introduce \textbf{\textsc{karma}}, a ViT-MAE variant that integrates the Linear Spectral Mixing Model (LSMM) as a physics-guided branch and enforces abundance constraints in an end-to-end manner.
  \item We design a hybrid objective combining Huber loss, Spectral Angle Mapper (SAM), and a physics-consistency term that jointly improves spectral fidelity and physical realism.
  \item We demonstrate the effectiveness of \textsc{karma} on the EnMAP hyperspectral benchmark, showing that physics-guided learning improves reconstruction fidelity, interpretability, and downstream transfer across both regression and classification tasks.
\end{itemize}
\section{Methodology}
\label{sec:method}

\begin{figure*}[!htbp]
  \centering
    \includegraphics[width=0.98\textwidth, alt={Proposed architecture diagram}]{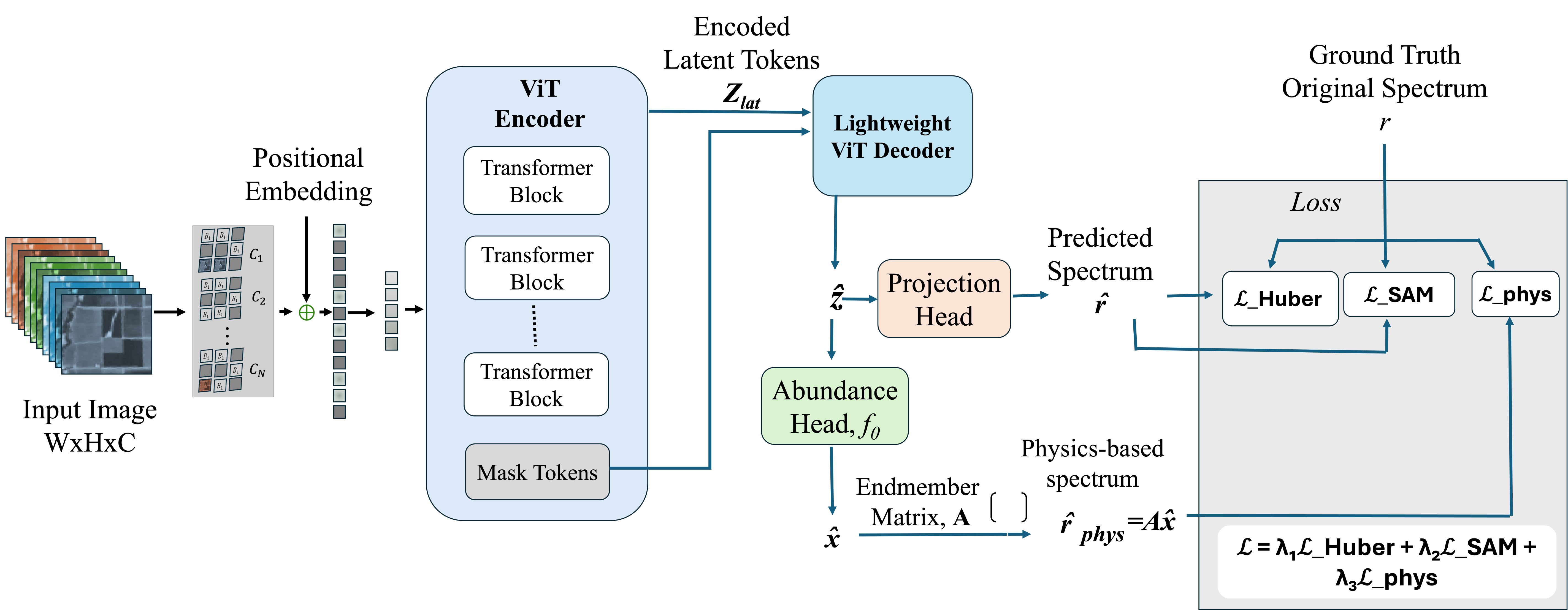}
  \caption{Architecture of \textsc{karma}: A knowledge-guided ViT-MAE that integrates the Linear Spectral Mixing Model (LSMM) within the decoder. The model jointly optimizes Huber loss, SAM, and physics-consistency losses to enhance spectral fidelity and interpretability.}
  \label{fig:prop_architecture}
  \vskip -0.15in
\end{figure*}

This section presents the proposed \textsc{karma} framework. We begin with an overview of the ViT-MAE backbone architecture, then elaborate on our physics-informed learning with LSMM, spectral constraint, SAM and describe the hybrid optimization objective. The complete proposed architecture is depicted in Figure~\ref{fig:prop_architecture}.

\subsection{ViT-MAE backbone}
The base architecture of \textsc{karma} builds upon the asymmetric Vision Transformer-based Masked Autoencoder (ViT-MAE) \cite{he2022masked,dosovitskiy2021vit}. The input hyperspectral cube is divided into non-overlapping patches, where each of the patches are flattened and linearly projected into an embedding vector. Following standard practice, a high masking ratio (typically 75\%) is applied to randomly mask a large portion of these patches, creating a reconstruction-based self-supervised task that encourages the model to learn robust contextual dependencies.

The encoder processes only the visible (unmasked) patches, producing compact latent representations that summarize spatial-spectral context. The lightweight decoder then reconstructs the masked patches using both the encoded latent tokens and learnable mask tokens. Unlike standard MAE architectures that predict pixel values directly, our decoder incorporates a physics-guided branch that explicitly models the physical composition of each spectral observation, introducing domain knowledge directly into the learning process.

\subsection{Integration of Linear Spectral Mixing Model (LSMM)}
To bridge physical modeling with transformer-based learning, we embed the LSMM into the decoder as a parallel reconstruction branch. LSMM is a physics-based model in remote sensing and spectral analysis which is used to describe how the spectral signature recorded by a sensor over a pixel is formed by the combination of different surface materials (endmembers) \cite{wei2020overview}. LSMM assumes that an observed signal can be represented as a linear combination of those latent components or \textit{endmembers}:

\begin{equation}
\mathbf{r} = \mathbf{A}\mathbf{x} + \mathbf{e},
\end{equation}
where $\mathbf{r} \in \mathbb{R}^{C}$ is the observed spectral vector of EnMap hyperspectral images with $C=218$ bands, $\mathbf{A} \in \mathbb{R}^{C \times M}$ is the endmember matrix containing the spectral signatures of $M$ constituent materials, $\mathbf{x} \in \mathbb{R}^{M}$ is the abundance vector, and $\mathbf{e}$ denotes reconstruction noise \cite{bioucas2012overview}. 

Physical validity requires that abundances satisfy two fundamental constraints,  \emph{sum-to-one} and \emph{non-negativity}:
\begin{equation}
\mathbf{x} \ge 0, \quad \mathbf{1}^\top \mathbf{x} = 1,
\end{equation}
These constraints ensure that the material mixtures remain physically meaningful. In our \textsc{karma} architecture, the decoder incorporates a compact \textit{abundance head} $f_\theta(\cdot)$, implemented as a lightweight multilayer perceptron (MLP). This module transforms each decoder patch token $\mathbf{z}$ into predicted abundance values. By applying softmax activation, we obtain
\begin{equation}
\hat{\mathbf{x}} = \operatorname{softmax}(f_\theta(\mathbf{z})),
\end{equation}
which inherently enforces both the non-negativity and simplex constraints without the need for additional regularization.

The LSMM-based reconstruction is then computed as:
\begin{equation}
\hat{\mathbf{r}}_{\text{phys}} = \mathbf{A}\hat{\mathbf{x}}.
\end{equation}
The endmember matrix $\mathbf{A}$ can be initialized from prior knowledge (e.g., spectral libraries or \textit{k}-Means prototypes) or learned end-to-end with a small learning rate for adaptive refinement \cite{ozkan2019endnet}. 

In this work, the endmember matrix $A$ is initialized randomly and learned end-to-end. In the context of the Masked Autoencoder, $A$ serves primarily as a set of \textbf{learned spectral basis vectors} rather than a fixed set of pure physical endmembers. By forcing the reconstruction $\hat{r}_{phys}=A\hat{x}$ through a linear combination of a small number of latent components $M$, the LSMM component acts as a powerful \textbf{low-rank physics bottleneck}. This compels the network to find an efficient, physically plausible, and stable decomposition of the data for effective reconstruction, enhancing the robustness of the learned features.

\subsection{Spectral Angle Mapper (SAM) for geometric alignment}
The Spectral Angle Mapper (SAM) offers a complementary perspective to traditional pixel-wise metrics by measuring the angular distance between spectral vectors, independent of their magnitude \cite{kruse1993sips,yuhas1992sam}. Unlike MSE or Huber Loss, which is sensitive to intensity variations, SAM focuses on preserving the spectral \emph{shape}, a critical property for identifying materials in hyperspectral data. We define the SAM loss across $N$ samples as:

\begin{equation}
\mathcal{L}_{\text{SAM}} = \frac{1}{N}\sum_{i=1}^{N} 
\arccos\!\left(
\frac{\langle \hat{\mathbf{r}}_i, \mathbf{r}_i \rangle}{
\|\hat{\mathbf{r}}_i\|_2 \, \|\mathbf{r}_i\|_2 + \epsilon}
\right),
\end{equation}
where $\langle \cdot , \cdot \rangle$ denotes the dot product, and $\epsilon$ is a small constant that prevents numerical instability from division by zero.

Minimizing $\mathcal{L}_{\text{SAM}}$ guides the reconstructed spectra $\hat{\mathbf{r}}$ to align directionally with the ground truth $\mathbf{r}$ in spectral space, even when absolute magnitudes differ. This geometric alignment helps the model maintain consistent representations and enhances its ability to discriminate between spectrally similar materials.

\subsection{Hybrid objective function}
\textsc{karma} optimizes three complementary objectives that jointly enforce numerical accuracy, spectral shape fidelity, and physical consistency. The unified loss function combines:

\begin{equation}
\mathcal{L} = \lambda_1 \mathcal{L}_{\text{Huber}} + 
\lambda_2 \mathcal{L}_{\text{SAM}} + 
\lambda_3 \mathcal{L}_{\text{phys}},
\end{equation}
where $\lambda_i$ are hyperparameters that denote the weights to control the contribution of each term.

\paragraph{(1) Huber Loss (Robust Regression).}
We employ the \textit{Huber loss} to improve robustness against spectral outliers and noise. 
The Huber loss behaves quadratically for small errors and linearly for large deviations, thereby balancing sensitivity and stability during reconstruction: 

\begin{equation}
\mathcal{L}_{\text{Huber}} = 
\frac{1}{N}\sum_{i=1}^{N} 
\begin{cases}
\frac{1}{2}\|\hat{\mathbf{r}}_i - \mathbf{r}_i\|_2^2, & 
\text{if } \|\hat{\mathbf{r}}_i - \mathbf{r}_i\|_2 \le \delta, \\
\delta\|\hat{\mathbf{r}}_i - \mathbf{r}_i\|_2 - \frac{1}{2}\delta^2, & 
\text{otherwise},
\end{cases}
\end{equation}

\paragraph{(2) Spectral Angle Mapper (SAM).}  
Enforces angular similarity and shape preservation, as defined above.

\paragraph{(3) Physics Consistency Loss.}  
Ensures reconstruction fidelity constrianed by LSMM:
\begin{equation}
\mathcal{L}_{\text{phys}} = 
\frac{1}{N}\sum_{i=1}^{N}
\|\mathbf{r}_i - \mathbf{A}\hat{\mathbf{x}}_i\|_2^2.
\end{equation}

Together, these losses create a balance between data-driven and knowledge-driven supervision, allowing the model to learn both expressive and interpretable representations.

\subsection{Architectural details}

\begin{figure}[!htbp]
  \centering
    \includegraphics[width=0.4\textwidth, height=4.5cm, alt={Ablation on Endmember head size}]{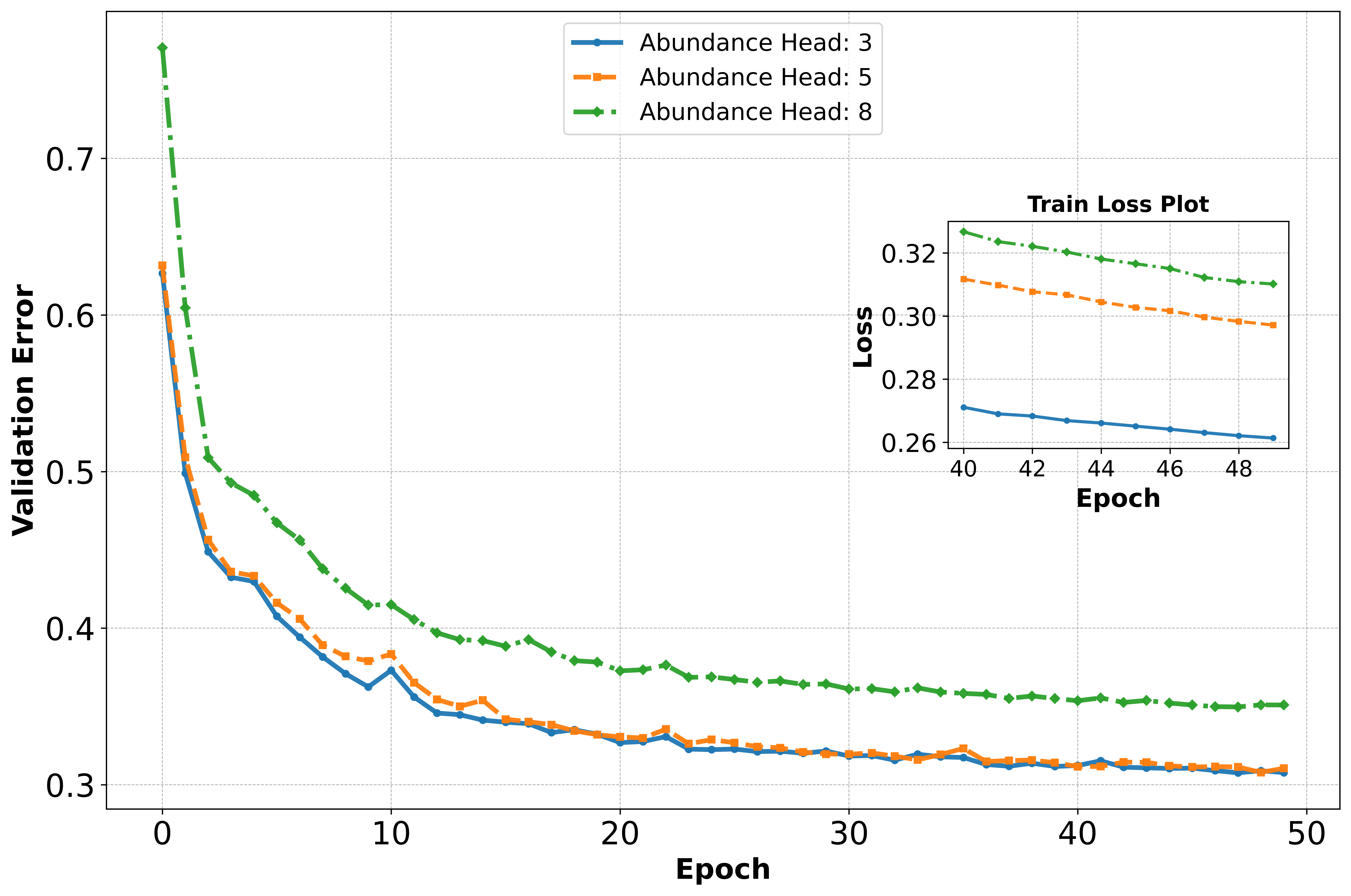}
  \caption{Ablation study on the number of abundance head components.}
  \label{fig:ablation-on-abundance-head}
  \vskip -0.1in
\end{figure}

Our architecture begins with patch embedding that projects the $C=218$ input channels (from the EnMap hyperspectral satellite images) into an embedding dimension $D$ (typically $D=512$). The multi-head self-attention mechanism employs $H$ heads with $D$ divisible by $H$ (commonly $H=8$). The abundance head comprises a compact MLP with architecture $D \rightarrow D/2 \rightarrow M$, producing abundance predictions for each patch token. We set the endmember matrix $\mathbf{A}$ dimensions to $218 \times M$, where $M$ typically based on dataset characteristics and ablation studies like as shown in Figure~\ref{fig:ablation-on-abundance-head}


 This physics-guided training regulates \textsc{karma} to ensure that reconstructions are both numerically accurate and physically consistent. This is an essential step toward interpretable foundation models for hyperspectral data.


\section{Experimental Setup}
\label{sec:experiments}

\subsection{Dataset and preprocessing}
\label{sec:dataset}

We conduct our experiments using hyperspectral imagery from the Environmental Mapping and Analysis Program (EnMAP)~\cite{storch2023enmap}, a spaceborne sensor that delivers observations across 224 spectral bands spanning the 420--2450~nm frequency range at 30~m spatial resolution. Our primary study region covers the state of California, which offers rich diversity in land-cover types ranging from dense forests and urban areas to heterogeneous agricultural landscapes. To test the generalization of our  model across different environmental conditions, we expand our downstream analysis to include representative areas in Colorado and Kansas.

Following radiometric calibration and atmospheric correction, we removed six bands affected by noise and retained 218 spectral bands for our analysis. We spatially resample the imagery into non-overlapping $224\times224$ pixel tiles, matching the input requirements of our ViT-MAE-based architecture. The dataset is organized into three subsets: 5,000 tiles for pretraining, 500 for validation, 2,000 for the primary test set (\textit{Test Dataset~1}). We conducted certain ablation studies on smaller subsets of these data due to computational constraints.



\subsection{Downstream Model}
\label{sec:downstream}

\begin{figure}[!htbp]
  \centering
    \includegraphics[width=0.48\textwidth, alt={Downstream Model}]{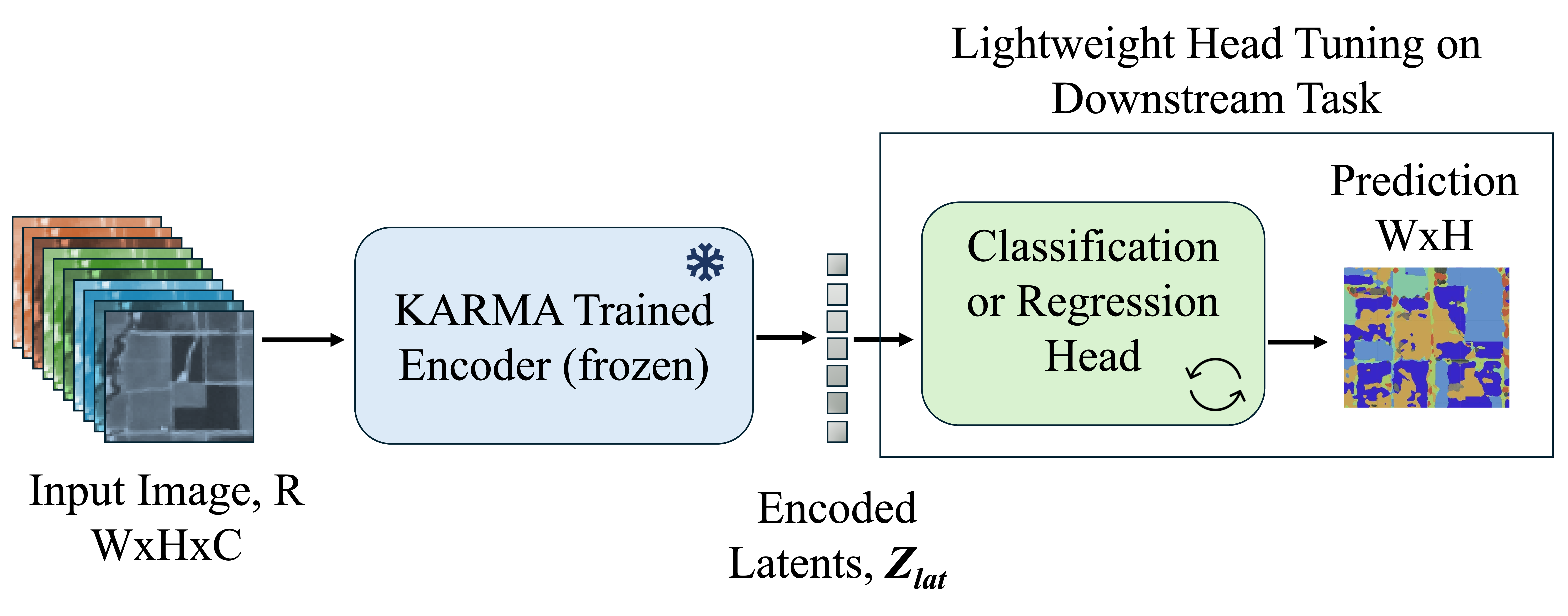}
  \caption{Downstream model consisting of the frozen \textsc{karma} encoder (pre-trained with KG) and a lightweight CNN head}
  \label{fig:downstream-framework}
  \vskip -0.15in
\end{figure}

As illustrated in Figure~\ref{fig:downstream-framework}, we leverage the pre-trained \textbf{\textsc{karma} encoder} to extract rich spectral–spatial representations from the 218-band EnMAP hyperspectral data. For downstream tasks, lightweight task-specific heads are attached to the frozen encoder to ensure efficient fine-tuning while preserving the learned representations. Here, we employ a compact classification head composed of a few convolutional and fully connected layers, optimized using the cross-entropy loss for two target applications.

\subsection{Training details}
We adopt the following default configuration: 16×16 patch size, embedding dimension $D=512$, $H=8$ attention heads, 75\% masking ratio, and batch size of 32. Training uses the AdamW optimizer with a base learning rate of $1\times10^{-4}$ and cosine decay schedule. Pretraining runs for 300-500 epochs depending on computational resources. 
The hybrid loss weight parameters,  $\lambda_1, \lambda_2, and \lambda_3$ are also chosen empirically. 

\subsection{Evaluation protocol}
Our evaluation follows a two-phase structure designed to assess both reconstruction quality and transferability of learned representations.

\textbf{Phase I: Reconstruction Quality.} 
We evaluate reconstruction performance using PSNR and SSIM. While PSNR assesses pixel-level fidelity, indicating how closely the reconstructed spectra match the reference, SSIM provides a perceptual measure of structural integrity, reflecting the model’s ability to preserve spatial and textural patterns. Furthermore, we explicitly compare KARMA against the \textbf{vanilla ViT-MAE} baseline to directly quantify the incremental benefit of integrating the LSMM and SAM  within the Masked Autoencoder architecture. While many stronger HSI-SOTA methods exist, a direct architectural comparison against ViT-MAE serves as the most essential ablation to prove that our knowledge-guided components are effective regularizers for this specific transformer framework.


\textbf{Phase II: Downstream Transfer Learning.} 
We utilize the pretrained \textsc{karma} encoder across two representative downstream tasks to assess the quality and generalization of its learned representations. The first task, crop type identification, is formulated as a multi-class classification problem spanning diverse agricultural categories. The second task, national land cover classification, captures broader spatial and ecological variability. To evaluate performance, we report both Top-1 Accuracy and Mean Intersection over Union (mIoU), which together measure categorical precision and spatial consistency in the model’s predictions.

\section{Results and Analysis}
\label{sec:results}
\begin{table*}[!htbp]
    \centering
    \setlength{\tabcolsep}{4pt} 
    \renewcommand{\arraystretch}{1.2} 
    \begin{tabular}{@{}clcc*{4}{c}@{}}
    \toprule
    \textbf{ID} & \textbf{Model} & \makecell{\textbf{Reconstruction} \\ \textbf{Loss}} & \makecell{\textbf{Physical} \\ \textbf{Constraints}} & \makecell{\textbf{Avg.} \\ \textbf{PSNR} ($\uparrow$)} & \makecell{\textbf{Per-Channel} \\ \textbf{PSNR (Max)} ($\uparrow$)} & \makecell{\textbf{Avg.} \\ \textbf{SSIM} ($\uparrow$)} & \makecell{\textbf{Per-Channel} \\ \textbf{SSIM (Max)} ($\uparrow$)} \\
    \midrule
    1 & VitMAE & HUBER & - & 24.61 & 32.45 & 0.55 & 0.82 \\
    2 & \textbf{KARMA} (Ours) & HUBER & SAM + LSMM & \cellcolor{lightgray}27.38 & \cellcolor{lightgray}35.10 & \cellcolor{lightgray}0.68 & \cellcolor{lightgray}0.88 \\
    \midrule
    \rowcolor{lightblue}
    \multicolumn{4}{l}{\textbf{Improvement} (\% over ViTMAE)} & \textbf{11.26} & \textbf{8.17} & \textbf{23.64} & \textbf{7.32} \\
    \bottomrule
    \end{tabular}
    \caption{Reconstruction accuracy comparison between \textsc{karma} and ViTMAE on hyperspectral data. The physics-guided learning in \textsc{karma} consistently yields superior reconstruction fidelity.}
    \label{tab:model_performance}
\end{table*}
\subsection{Reconstruction Fidelity Analysis}
Table~\ref{tab:model_performance} demonstrates that incorporating physical knowledge through the Linear Spectral Mixing Model (LSMM) and Spectral Angle Mapper (SAM) loss improves reconstruction quality over the purely data-driven baseline. \textsc{karma} achieves an average PSNR of 27.38~dB compared to ViTMAE's 24.61~dB, representing a 2.77~dB improvement. This corresponds to an 11.3\% relative gain in reconstruction accuracy.

The improvement is more pronounced when measured by the Structural Similarity Index Measure (SSIM), which captures structural pattern preservation beyond pixel-wise differences. \textsc{karma} achieves an average SSIM of 0.68 compared to ViTMAE's 0.55, a relative improvement of 23.6\%. The higher SSIM scores indicate that \textsc{karma} produces reconstructions that preserve physically meaningful spectral-spatial structures, validating the incorporation of domain-specific physical constraints into the masked autoencoder framework.
\subsection{Computational Overhead}
We empirically measured the average training time per sample (Table \ref{tab:timing}) to quantify the computational cost added by the knowledge-guided components. The incorporation of the \textbf{SAM} introduces an overhead of approximately $26\%$ compared to the ViT-MAE baseline. The full \textsc{KARMA} model increases the time to $9.47$ ms, representing an overall overhead of approximately $31.7\%$. While this is a measurable cost only for training, and \textbf{sublinear} relative to the substantial performance gains achieved. 

\begin{table}[h] 
    \centering
    \setlength{\tabcolsep}{2pt} 
    \renewcommand{\arraystretch}{1.15} 
    \begin{tabular}{@{}l c @{}}
    \toprule
    \textbf{Model} & \textbf{Avg. Time per sample (ms)} \\
    \midrule
    Baseline ViTMAE & $7.19$  \\
    ViTMAE+SAM & $9.06$  \\
    \textbf{KARMA (SAM+LSMM)} & $\mathbf{9.47}$ \\
    \bottomrule
    \end{tabular}
        \caption{Computational trade-off for Physics-Informed training on single RTX 4000 GPU with batch size 16.}
    \label{tab:timing}
\end{table}

\subsection{Downstream Task Performance and Generalization}

While the classification of certain classes in CDL and NLCD is inherently challenging due to noisy labels and dataset preparation issues, the improvements in self-supervised reconstruction translate into significant gains across downstream tasks. This highlights the interpretability and transferability of \textsc{karma}'s learned features.

\begin{table}[!htbp]
    \centering
    \setlength{\tabcolsep}{2pt} 
    \renewcommand{\arraystretch}{1.15} 
    
    \begin{tabular}{@{}l c c c c@{}}
    \toprule
    \textbf{Class Name} & \multicolumn{2}{c}{\textbf{VitMAE + Head}} & \multicolumn{2}{c}{\textbf{KARMA + Head}} \\
    \cmidrule(lr){2-3} \cmidrule(lr){4-5}
    & \makecell{Top-1\\Acc(\%) ($\uparrow$)} & \makecell{mIoU\\(\%) ($\uparrow$)} & \makecell{Top-1\\Acc(\%) ($\uparrow$)} & \makecell{mIoU\\(\%) ($\uparrow$)} \\
    \midrule
    Shrubland & 40.31 & 34.74 & \cellcolor{lightgray}71.18 & \cellcolor{lightgray}51.35 \\
    Grassland/Pasture & 22.74 & 16.85 & \cellcolor{lightgray}44.22 & \cellcolor{lightgray}31.47 \\
    Evergreen Forest & 81.72 & 53.04 & \cellcolor{lightgray}85.03 & \cellcolor{lightgray}56.30 \\
    \midrule 
    \rowcolor{lightblue}
    \textbf{Average} & 48.26 & 34.88 & \textbf{66.81} & \textbf{46.37} \\
    \bottomrule
    \end{tabular}
        \caption{Crop Type Identification on CDL data (top three classes in California). \textsc{karma} with a lightweight CNN significantly outperforms the baseline.}
    \label{tab:cdl_performance}
    \end{table}

\subsubsection{Crop Type Identification (CDL Data)}

As reported in Table \ref{tab:cdl_performance}, \textsc{karma} paired with a lightweight classification head substantially outperforms ViTMAE. On average, Top-1 Accuracy rises from 48.25\% to 63.12\% (a 30.8\% relative improvement), and mean IoU increases from 35.67\% to 50.88\% (a 42.6\% relative gain). The largest improvements are observed in spectrally complex and mixed classes, demonstrating that LSMM-guided pre-training produces discriminative and physically grounded latent features despite dataset noise.

\subsubsection{Land Cover Classification (NLCD Data)}
Table \ref{tab:nlcd_performance} evaluates generalization by testing the \textsc{karma} encoder, trained on California data, on geographically distinct regions (Colorado and Kansas). \textsc{karma} consistently achieves higher accuracy across all classes, with the most notable improvement in "Cultivated Crops," rising from 56.70\% to 91.59\% (a 61.5\% relative increase). Overall, average Top-1 Accuracy improves from 61.24\% to 79.83\% (+30.3\% relative) and mean IoU from 44.18\% to 63.47\% (+43.6\% relative). These results confirm that \textsc{karma} captures physics-based inductive biases, yielding features that are robust and transferable across noisy datasets and diverse terrains.



\begin{table}[!http]
    \centering
    \setlength{\tabcolsep}{6pt} 
    \renewcommand{\arraystretch}{1.15} 
    
    \begin{tabular}{@{}l c c@{}}
    \toprule
    \textbf{Class Name} & \multicolumn{2}{c}{\textbf{Top-1 Accuracy (\%)} ($\uparrow$)} \\
    \cmidrule(lr){2-3}
    & ViTMAE + Head & \textbf{KARMA} + Head \\
    \midrule
    Shrub/Scrub & 63.18 & \cellcolor{lightgray}70.50 \\
    Herbaceous & 35.43 & \cellcolor{lightgray}46.59\\
    Cultivated Crops & 56.70 & \cellcolor{lightgray}91.59 \\
    \midrule 
    \rowcolor{lightblue}
    \textbf{Average} & 51.77 & \textbf{69.56} \\
    \bottomrule
    \end{tabular}
        \caption{Land Cover Classification on NLCD data (top three classes across CA, CO, and KS). The \textsc{karma} encoder, trained solely on California data, generalizes effectively to other regions (Colorado and Kansas), demonstrating improved downstream performance.}
    \label{tab:nlcd_performance}
\end{table}

\section{Related Work}
\label{sec:related}

\paragraph{Self-supervised vision and masked modeling.}
MAEs and related masked modeling approaches have rapidly advanced representation learning for vision, showing strong transfer to downstream tasks \cite{he2022masked,dosovitskiy2021vit}. Extensions to multi-spectral and temporal remote sensing (e.g., SatMAE) demonstrate benefits from spectral-temporal pretext tasks \cite{cong2022satmae}. Building on this direction \textit{HyperKD}~\cite{matin2025hyperkd}, extended the ViT-MAE paradigm to cross-spectral settings through inverse domain knowledge distillation, effectively bridging multispectral and hyperspectral modalities. 
In contrast, the present study departs from distillation-based strategies and instead introduces a physics-guided mechanism that embeds the Linear Spectral Mixing Model (LSMM) directly within the ViT-MAE architecture to enhance reconstruction fidelity and interpretability.

\paragraph{Physics-informed and knowledge-guided learning.}
Integrating scientific priors into ML models has been formalized under physics-informed learning and knowledge-guided ML; these approaches provide improved physical realism and extrapolation \cite{raissi2019pinn,willard2020integrating,karpatne2024kgmlsurvey,liu2024kgml}. KGML methods have been used to incorporate support for physical constraints grounded in soil hydrology, including the van Genuchten water retention equations and models (i.e., Richards’ Equation) of hydraulic conductivity \cite{khandelwal2024deepsoil, khandelwal2025subterra}; vegetation indices \cite{bruhwiler2020lightweight}; evapotranspiration \cite{armstrong2022attention}; preserving graph properties such as betweenness centrality \cite{matin2022rapid}; masking cloud occlusions in satellite imagery \cite{khandelwal2022cloudnet}, accounting for human perceptual limits during visualization \cite{mitra2021glance}; masked autoencoders \cite{faruk2025accounting, matin2025hyperkd} and accounting for correlations between soil spectroscopic properties \cite{bachinin2025science, dey2025deepsalt}.  Our \textsc{karma} framework follows this paradigm by embedding LSMM within a self-supervised transformer framework. Finally, knowledge distillation approaches have been explored in the context of physical phenomena \cite{matin2023discern}; our methodology is complementary to these approaches.

\paragraph{Spectral mixing and unmixing.}
LSMM and hyperspectral unmixing have a rich literature; classical formulations and modern deep unmixing approaches show how material signatures can be recovered and used for interpretability \cite{bioucas2012overview,ozkan2019endnet}. \textsc{karma} uses LSMM as an explicit constraint rather than a standalone unmixing network.

\section{Conclusion and Future Work}
\label{sec:conclusion}
We introduced \textsc{karma}, a knowledge-guided masked autoencoder that bridges data-driven learning with physical principles for hyperspectral image analysis. By integrating the Linear Spectral Mixing Model into the decoder and using a hybrid loss that combines Huber, SAM, and physics-based constraints, our approach yields more accurate and physically interpretable reconstructions. Our experiments with EnMAP hyperspectral imagery demonstrates that this physics-guided self-supervision strategy not only improves reconstruction fidelity but also yields representations that transfer more effectively to downstream tasks. Our work adds to the growing body of evidence showing that incorporating domain theoretic knowledge into modern self-supervised architectures leads to models that achieve strong performance while remaining scientifically meaningful. This is particularly crucial for remote sensing applications where physical consistency matters as much as predictive performance.

As we move forward, we plan to broaden our study evaluating \textsc{karma} against several established baselines. Complementary ablation studies will systematically analyze the contribution of each component—such as the SAM loss, LSMM integration, the role of perfect physical fixed versus learnable endmember matrices $A$, and the effect of varying the number of endmembers $M$. Beyond these evaluations, we plan to explore scalability and cross-domain generalization by combining hyperspectral data with complementary sensing modalities. Together, these directions will strengthen the bridge between physical understanding and machine learning, ultimately leading to more robust, reliable, and interpretable Earth observation models.

\section{Acknowledgments}
This research was supported by the National Science Foundation (1931363,2312319), the National Institute of Food Agriculture (COL014021223, 2025-77039-45531), and an NSF/NIFA Artificial Intelligence Institutes AI-LEAF Award [2023-03616].

\bibliography{aaai24}

\end{document}